\documentclass[letterpaper, 10 pt, conference]{ieeeconf}  

\IEEEoverridecommandlockouts                              

\title{\LARGE \bf
H3O: Hyper-Efficient 3D Occupancy Prediction with Heterogeneous Supervision
}

\author{Yunxiao Shi$^{\dagger}$, Hong Cai$^{\dagger}$, Amin Ansari$^{\ddagger}$, and Fatih Porikli$^{\dagger}$
\thanks{$^{\dagger}$Qualcomm AI Research, an initiative of Qualcomm Technologies, Inc. Email: \{yunxshi, hcai, fporikli\}@qti.qualcomm.com.}
\thanks{$^{\ddagger}$Qualcomm Technologies, Inc. Email: amina@qti.qualcomm.com.}
}

\usepackage{amsmath}
\usepackage{amssymb}
\usepackage{amsfonts}
\usepackage{booktabs}
\usepackage{bm}
\usepackage{float}
\usepackage{graphicx}
\usepackage{makecell}
\usepackage{xcolor}
\usepackage{rotating}
\usepackage[hidelinks=True]{hyperref}
\usepackage{textcomp}
\usepackage{arydshln}
\usepackage{xspace}
\usepackage{wasysym}
\usepackage{cite}

\newcommand{\ours}{{H3O}\xspace}
\newcommand{\ie}{\textit{i.e.}}
\newcommand{\eg}{\textit{e.g.}}

\begin{document}

\maketitle
\thispagestyle{empty}
\pagestyle{empty}

\begin{abstract}
3D occupancy prediction has recently emerged as a new paradigm for holistic 3D scene understanding and provides valuable information for downstream planning in autonomous driving. Most existing methods, however, are computationally expensive, requiring costly attention-based 2D-3D transformation and 3D feature processing. In this paper, we present a novel 3D occupancy prediction approach, \ours, which features highly efficient architecture designs that incur a significantly lower computational cost as compared to the current state-of-the-art methods. In addition, to compensate for the ambiguity in ground-truth 3D occupancy labels, we advocate leveraging auxiliary tasks to complement the direct 3D supervision. In particular, we integrate multi-camera depth estimation, semantic segmentation, and surface normal estimation via differentiable volume rendering, supervised by corresponding 2D labels that introduces rich and heterogeneous supervision signals. We conduct extensive experiments on the Occ3D-nuScenes and SemanticKITTI benchmarks that demonstrate the superiority of our proposed \ours. 


\end{abstract}

\section{Introduction}

Perceiving the 3D world from 2D images is critical for various important applications such as autonomous driving~\cite{hu2023planning} and augmented/virtual reality~\cite{grauman2022ego4d}. Traditional tasks like depth perception~\cite{godard2019digging,zhu1232020mda,shi2023ega,shi2024decotr,yasarla2023mamo,yasarla2024futuredepth} and 3D object objection~\cite{wang2021fcos3d,wang2022detr3d,jiang2024far3d,li2022bevformer,yang2023bevformer} have made significant progress in advancing vision-centric 3D perception. However, the per-pixel representation of depth and the coarse 3D bounding box presents a challenge towards a holistic understanding of fine-grained 3D scene geometry and semantics. 

Recently, 3D occupancy prediction has garnered significant interest from both industry and academia as a new paradigm for 3D scene understanding. Specifically, 3D occupancy prediction aims to infer fine-grained 3D geometry and semantics from camera images, providing the level of granularity that depth and 3D object detection does not possess, which is crucial for downstream tasks such as motion planning for autonomous driving.

Existing 3D occupancy solutions learn from ground-truth 3D occupancy labels which are derived from LiDAR scans. However, curating high-quality, dense 3D occupancy ground-truth is complicated and time-consuming~\cite{wang2023openoccupancy}. Even with latest (semi)-automated pipelines~\cite{wei2023surroundocc,tian2024occ3d}, the 3D labels can be ambiguous or incorrect in some areas, due to factors such as sparsity, occlusion, and misalignment~\cite{wei2023surroundocc,tian2024occ3d}. 
Several more recent works~\cite{huang2024selfocc,gan2024comprehensive} have explored self-supervision for 3D occupancy but the results still fall short of the supervised methods~\cite{huang2023tri,wei2023surroundocc}. 

Moreover, existing solutions are often computationally heavy and memory intensive leveraging costly operations like cross-attention to perform 2D-3D transformation that constructs the 3D feature volume~\cite{huang2023tri,wei2023surroundocc,zhang2021nerfactor}, which makes it challenging to deploy especially on resource constrained platforms with limited onboard compute such as autonomous vehicles.  

\begin{figure}[t!]
  \centering
  \includegraphics[width=0.38\textwidth]{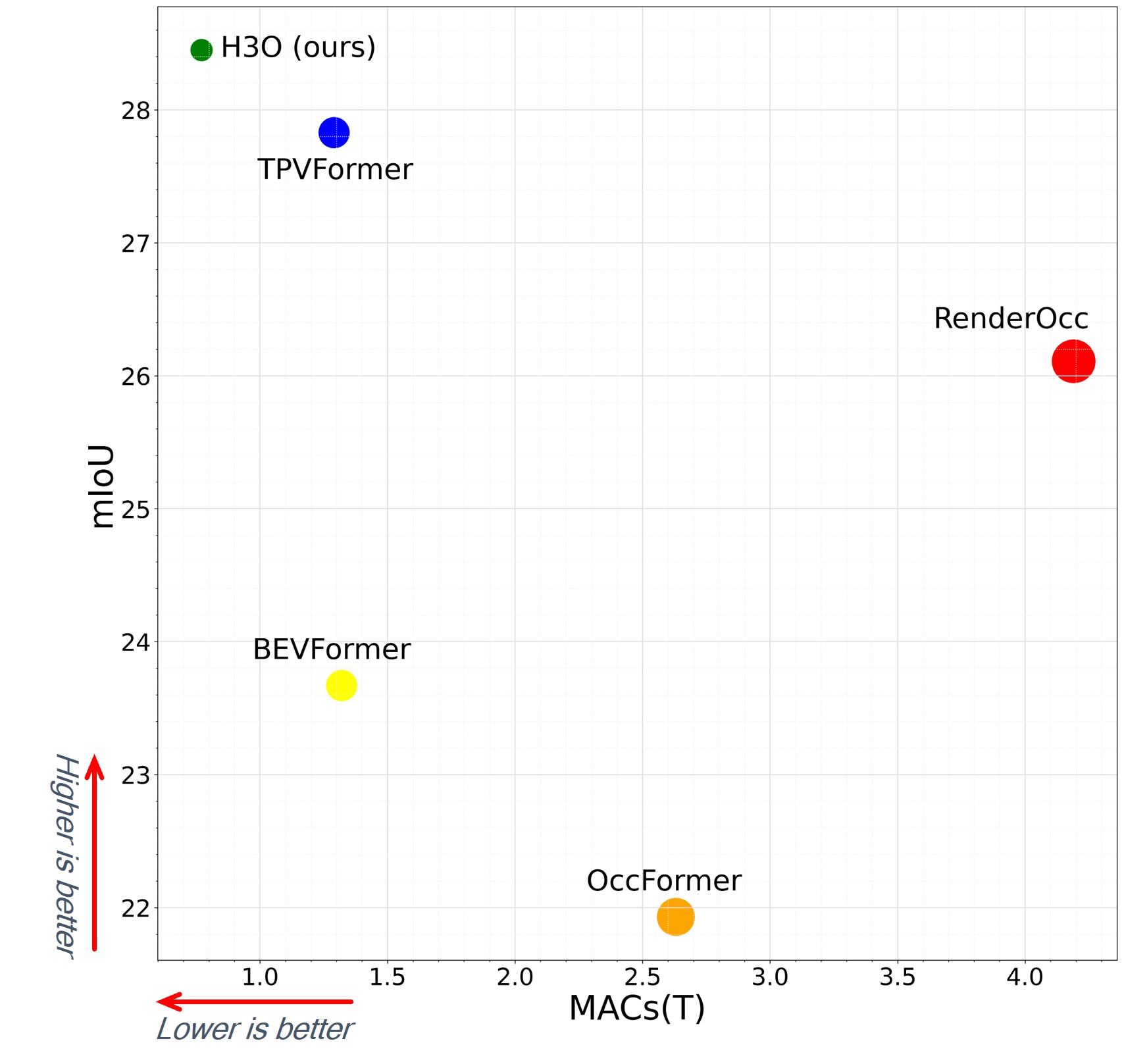}
  \vspace{-5pt}
  \caption{\small Accuracy (mIoU on Occ3D-nuScenes~\cite{tian2024occ3d}) vs. computation efficiency (Tera-MACs). Our proposed \ours achieves the best accuracy and efficiency when comparing to baseline and latest state-of-the-art methods, including BEVFormer~\cite{li2022bevformer}, TPVFormer~\cite{huang2023tri}, OccFormer~\cite{zhang2023occformer}, and RenderOcc~\cite{pan2024renderocc}. Note that in this work, we consider the setting where input multi-camera images are from the same time step, \ie, we do not consider video frames.}
  \label{fig:overview}
  \vspace{-10pt}
\end{figure}

\begin{figure*}[t!]
    \centering
    \includegraphics[width=0.98\textwidth]{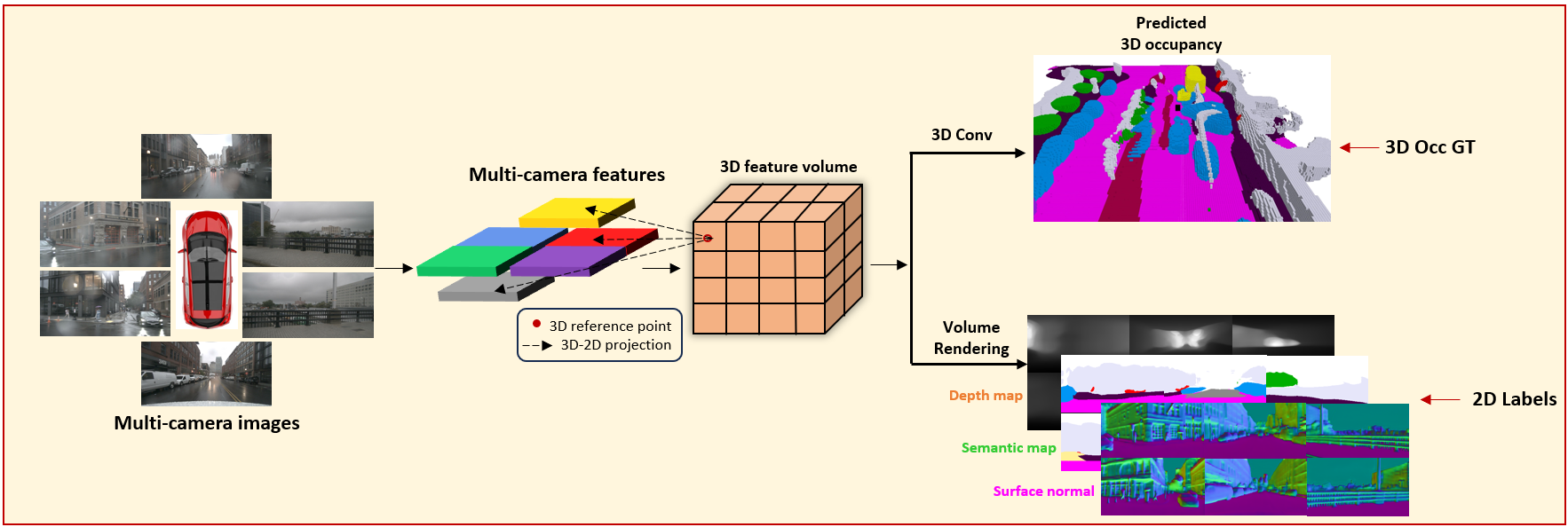}
    \caption{\small Overview of our proposed \ours. First, an image backbone (\eg, ResNet~\cite{he2016deep}) extracts features from the multiple camera images. Then, a 3D grid is used to query the 2D image features, based on which \ours constructs a 3D feature volume. Specifically, we average features from different views and avoid costly operations like cross-attention, which leads to significantly better efficiency. The 3D feature volume is subsequently processed by 3D convolutions to generate occupancy predictions. During training, volume rendering is also performed based on the 3D feature volume to produce 2D predictions, including depth, semantics, and surface normals, which are supervised to provide additional training signals. Note that volume rendering is disabled at inference-time hence incurring zero extra cost.}
    \label{fig:diagram}
    \vspace{-10pt}
\end{figure*}

In this paper, we propose a hyper efficient and performing 3D occupancy prediction approach, named \ours. Our approach first efficiently constructs a 3D volume as scene representation. We generate 3D reference points and project them to each view with known camera parameters, followed by bilinear interpolation to obtain image features per view after which we directly average them to obtain voxel features instead of the costly 2D-3D cross attention used in~\cite{huang2023tri,wei2023surroundocc}. 

To complement learning from 3D occupancy ground-truth, \ours integrates auxiliary tasks of heterogeneous nature. In particular, we advocate multi-camera depth estimation, semantic segmentation, and surface normal estimation as the auxiliary tasks. Through differentiable volume rendering, we render 2D depth maps, semantic maps, and normal maps from the constructed 3D volume for each camera view, which are supervised by available 2D ground-truth labels or off-the-shelf 2D foundation models like~\cite{hu2024metric3d,yin2023metric3d}. Since 3D ground-truth occupancy annotations can be ambiguous in certain regions, introducing such heterogeneous supervision benefits the 3D occupancy learning.

Our main contributions are summarized as follows.
\begin{itemize}
    \item We introduce an efficient and accurate 3D occupancy prediction approach, \ours, which advocates integrating heterogeneous auxiliary tasks of multi-camera depth estimation, semantic segmentation and surface normal estimation through differentiable volume rendering, supervised by available LiDAR labels and foundational models, to complement and improve 3D occupancy learning.
    \item We propose an efficient way to construct the 3D volume relying only on perspective projection and bilinear interpolation, without the need for costly attention-based 2D-3D transformations. A simple yet effectively per-pixel ray sampling strategy is also advocated.
    \item We perform extensive experiments and compare with existing methods on the Occ3D-nuScenes~\cite{tian2024occ3d} and SemanticKITTI~\cite{behley2019semantickitti} benchmarks. \ours achieves state-of-the-art performance while incurs significantly less computation cost, opening doors for deployment on resource constrained platforms.
\end{itemize}


\section{Related Works}

\subsection{3D Occupancy Prediction}

The scalability and economic benefit of vision-based 3D perception have shown increasing promise for autonomous systems. 3D semantic occupancy prediction has garnered significant interest as an attractive alternative to LiDAR perception. It represents the scene using a discrete voxel grid, and estimates the occupancy state of each voxel and assigns semantic labels to occupied voxels. A number of approaches have been proposed to effectively learn 3D occupancy from images. MonoScene~\cite{cao2022monoscene} proposes a monocular 3D semantic occupancy system which employs 3D UNet to process voxels generated by sight projection. BEVFormer~\cite{li2022bevformer} describes the scene using a BEV grid generated from image features and leverages cross-attention to refine the grid. TPVFormer~\cite{huang2023tri} proposes a tri-perspective view as scene representation to learn 3D occupancy. SurroundOcc~\cite{wei2023surroundocc} designs an automated pipeline to generate dense occupancy ground truth to train occupancy networks in lieu of sparse LiDAR points used in~\cite{huang2023tri}. RenderOcc~\cite{pan2024renderocc} advocates the idea of using 2D labels to train 3D occupancy networks to reduce reliance on 3D occupancy annotations. Several works~\cite{cao2023scenerf,huang2024selfocc} explore self-supervised learning based on the photometric consistency between neighbouring frames to learn 3D occupancy. Meanwhile, several benchmarks~\cite{behley2019semantickitti,wei2023surroundocc,tian2024occ3d,tong2023scene,openscene2023} have been established to provide 3D occupancy ground truth based on existing datasets~\cite{geiger2012we,geiger2013vision,caesar2020nuscenes,sun2020scalability}. 

\subsection{Neural Radiance Fields and 3D Reconstruction}
Neural Radiance Field (NeRF) has witnessed explosive growth since the seminar work~\cite{mildenhall2021nerf}. It enables explicit or implicit control of scene properties such as illumination, camera parameters, pose, geometry, appearance, and semantic structure~\cite{tewari2020state}. NeRF~\cite{mildenhall2021nerf} learns the scene geometry by optimizing a continuous scene function over a collection of posed images. Due to its slow rendering speed and only being able to handle static scenes, subsequent works have dedicated on improving the efficiency of NeRF and making it able to render dynamic scenes~\cite{zhang2020nerf++,pumarola2021d,fridovich2022plenoxels,sun2022direct,muller2022instant}. Different from 3D occupancy prediction, the per-object or scene optimization scheme of NeRF-like methods emphasize more on rendering quality and cannot generalize across scenes. Yet, the technique of differentiable volume rendering is insightful and can be used to facilitate multi-task learning which can benefit 3D occupancy prediction, as we will show in this paper.

\section{Method}

\subsection{Problem Definition} 

3D occupancy prediction aims to produce a dense semantic voxel grid from multi-camera images which capture the surrounding environment. Given an ego-vehicle at time $t$, the system takes $N_C$ camera images, $\mathbf{I}=\{I_i\}_{i=1}^{N_C}$, as input and predicts the 3D semantic occupancy volume $\mathbf{O}\in\mathbf{R}^{C\times H\times W\times Z}$,
where $H,W,Z$ denotes the resolution of the volume and $C$ is the number of classes. We can formally describe 3D occupancy prediction as follows:
\begin{equation}
    \mathbf{O}=G(\mathbf{V}),\quad \mathbf{V}=F(\mathbf{I}),
\end{equation}
where $F(\cdot)$ consists of the image backbone that extracts multi-camera features and transforms them to 3D volume features $\mathbf{V}$, and $G(\cdot)$ is another neural network that maps $\mathbf{V}$ to occupancy predictions.

\subsection{Architecture Overview}

The overall pipeline of \ours is shown in Fig.~\ref{fig:diagram}. \ours first takes multi-camera images as input and extracts multi-camera image features through an image backbone. Next, to construct the 3D feature volume, we first generate the 3D points $\mathbf{P}=\{P=(p_x, p_y, p_z)\}$ using a reference voxel grid, and then project them back to each camera view $c=\{1, ..., N_C\}$, followed by bilinear interpolation to sample the 2D image features as follows:
\begin{equation}
    \mathcal{F}^{inter}_c = \mathcal{F}_c\langle\pi(\mathbf{P}, T, K)\rangle,
\end{equation}
where $\pi(\cdot)$ is the projection that maps the 3D point $P$ to the image plane. $T, K$ are the camera extrinsics and intrinsics, respectively. $\langle\cdot\rangle$ is the bilinear interpolation operator. $\mathcal{F}_c$ are the extracted image features and $\mathcal{F}^{inter}_c$ are the interpolated features.

Different from existing works~\cite{li2022bevformer,huang2023tri,wei2023surroundocc},  which predominately leverage costly cross-attention to aggregate features, we simply average all of the 2D camera features $F_c^{inter}$ from different views to obtain the voxel feature volume,
\begin{equation}
    \mathbf{V} = \frac{1}{N}\sum_{c=1}^{N_C}\mathcal{F}_c^{inter},
\end{equation}
which is a significantly more efficient operation and we find it sufficiently effective for constructing the 3D feature volume for the occupancy prediction task.

To enable features interaction across neighboring voxels, we use 3D convolutions to process $V$ and generate the final occupancy predictions.

\subsection{Volume Rendering and Heterogeneous Supervision}
Inspired by recent works~\cite{huang2024selfocc,pan2024renderocc,zhang2023occnerf}, we impose three auxiliary tasks, namely, multi-camera depth estimation, semantic segmentation,  and surface normal estimation, as shown Fig.~\ref{fig:diagram}, which helps enforce the multi-view consistency of learned occupancy.

To render depth, semantics and surface normals, We adopt differentiable volume rendering~\cite{mildenhall2021nerf,barron2021mip,verbin2022ref}. For rendering the depth of a pixel, a ray $\mathbf{r}$ from the camera center $\mathbf{o}$ is cast along the viewing direction $\mathbf{d}$ pointing to the pixel. Formally, $\mathbf{r}$ can be formulated as
\begin{equation}
    \mathbf{r}(t) = \text{o} + t\text{d},\text{ }t\in[t_s, t_e].
\end{equation}
We then sample $M$ points $\{t_i\}_{i=1}^M$ following $U[0, 1]$ along the ray  to get the density $\tau(t_i)$. Given the sampled $M$ points, the depth of the corresponding pixel is obtained by
\begin{equation}
    D^{pix} = \sum_{i=1}^MT(i)(1-\exp(-\tau(t_i)\delta_i)t_i,
\end{equation}
where $T(t_i) = \exp(-\sum_{j=1}^{i-1}\tau(t_i)\delta_i)$ and $\delta_i = t_{i+1} - t_i$ denotes the intervals between the sampled points.

To render 2D semantic maps, an additional semantic head with $C$ output channels is employed to map volume features $V$ to semantic outputs $S$. We make use of volume rendering again to get per-pixel semantic output
\begin{equation}
    S^{pix} = \sum_{i=1}^{M_s}T(t_i)(1-\exp(-\tau(t_i))\delta_i)S(t_i),
\end{equation}
where $M_s=\alpha M,\,\alpha\in(0, 1]$. 

To generate 2D surface normal maps, we use a spatial MLP to predict unit vectors at any 3D point location, which is a common approach in neural rendering~\cite{zhang2021nerfactor,bi2020neural}. Note that while one can derive normals from computing gradients of the volume density w.r.t. the 3D points~\cite{boss2021nerd,chen2021nerv,yu2022monosdf}, we adopt the MLP-based approach due to its simplicity.



\begin{figure}[t!]
  \centering
  \includegraphics[width=0.45\textwidth]{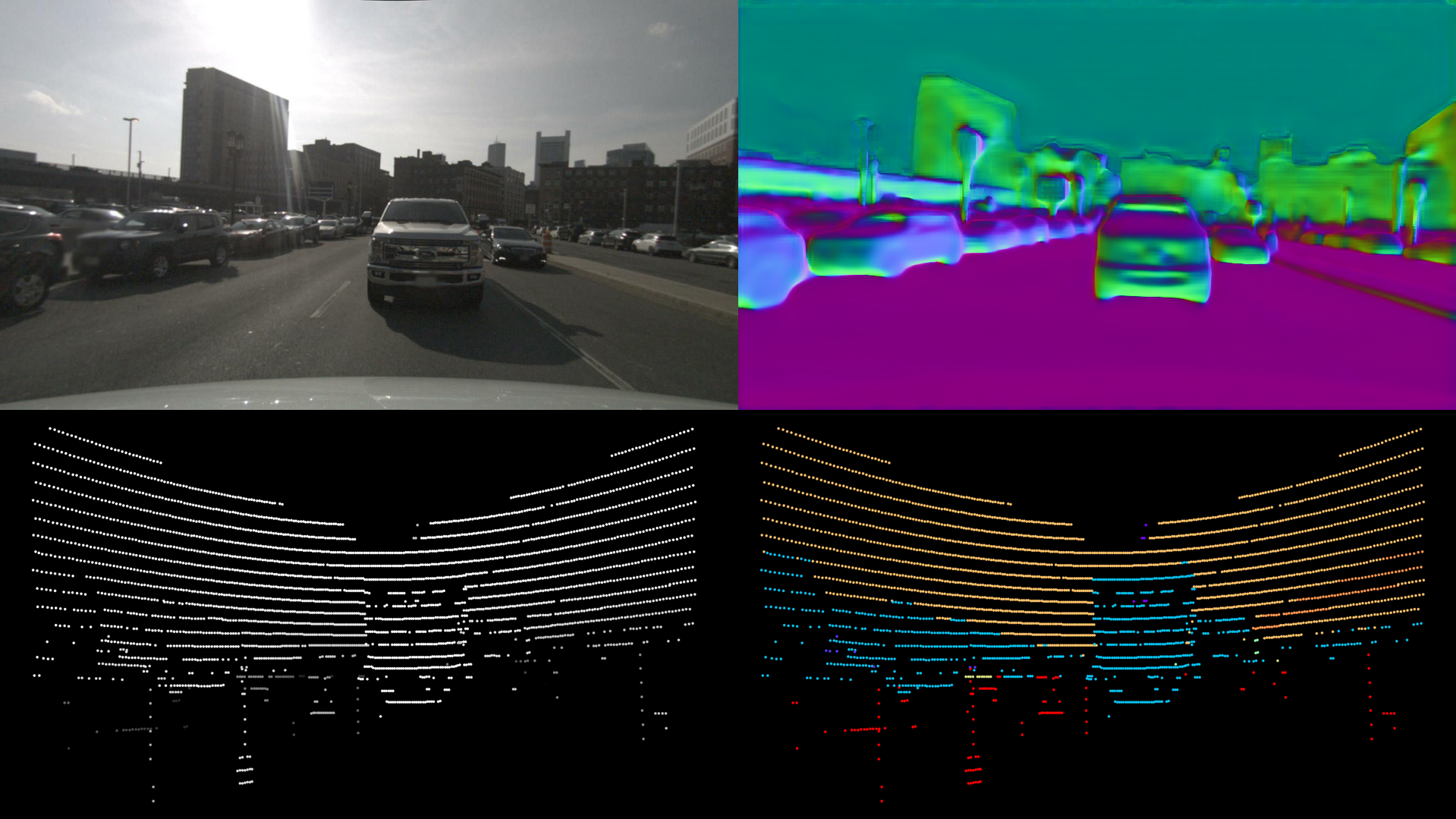}
  \caption{\small An example our multi-task supervision signals on nuScenes. Top row shows an RGB image and the normal map generated by Metric3Dv2~\cite{hu2024metric3d}, which we use as ground-truth normal. Bottom row shows the depth and semantic labels projected from LiDAR points. Best viewed when zoomed in.}
  \label{fig:2dlabel}
  \vspace{-10pt}
\end{figure}

To supervise rendered depth and semantics, we project LiDAR points to each camera view to obtain the corresponding 2D labels.\footnote{Semantic annotations on LiDAR point clouds are available in datasets like nuScenes~\cite{caesar2020nuscenes} and SemanticKITTI~\cite{behley2019semantickitti}.} For surface normal supervision, since there is no readily available ground-truth labels, we utilize recent powerful 2D foundation models~\cite{yin2023metric3d,hu2024metric3d} to generate normal maps for each camera view and use them as ground truth. Fig.~\ref{fig:2dlabel} shows an example of the auxiliary supervisions.

Note that due to its costly 3D volume, \cite{pan2024renderocc} requires an auxiliary-ray scheme in volume rendering in order to cover objects from different views; however, this introduces misalignment and noise in training. In contrast, thanks to our efficient design, we simply cast a ray for every pixel in the scene, covering all objects in the scene which yields more precise rendering.


\begin{table*}[t!]
  \small
  \centering
  \caption{\small 3D semantic occupancy results on Occ3D-nuScenes validation set~\cite{caesar2020nuscenes}.}
  \vspace{-5pt}
  \resizebox{\textwidth}{!}{
  \begin{tabular}{l|c|ccccccccccccccccc}
    \toprule
    \textbf{Method} & \makecell{SSC\\mIoU} & \makecell{\begin{turn}{90}others\end{turn}} & \makecell{\begin{turn}{90}barrier\end{turn}} & \makecell{\begin{turn}{90}bicycle\end{turn}} & \makecell{\begin{turn}{90}bus\end{turn}} & \makecell{\begin{turn}{90}car\end{turn}}& \makecell{\begin{turn}{90}cons. veh\end{turn}}& \makecell{\begin{turn}{90}motorcycle\end{turn}} & \makecell{\begin{turn}{90}pedestrian\end{turn}} & \makecell{\begin{turn}{90}traffic cone\end{turn}} & \makecell{\begin{turn}{90}trailer\end{turn}} & \makecell{\begin{turn}{90}truck\end{turn}} & \makecell{\begin{turn}{90}dri. sur\end{turn}} & \makecell{\begin{turn}{90}other flat\end{turn}}& \makecell{\begin{turn}{90}sidewalk\end{turn}} & \makecell{\begin{turn}{90}terrain\end{turn}} & \makecell{\begin{turn}{90}manmade \end{turn}} & \makecell{\begin{turn}{90}vegetation\end{turn}}\\
    \midrule
    MonoScene~\cite{cao2022monoscene} & 6.06 &1.75 &7.23 &4.26 &4.93 &9.38 &5.67 &3.98 &3.01 &5.90 &4.45 &7.17 &14.91 &6.32 &7.92 &7.43 &1.01 &7.65\\
    BEVFormer~\cite{li2022bevformer} & 23.67 &5.03 &38.79 &9.98 &34.41 &41.09 &13.24 &16.50 &18.15 &17.83 &18.66 &27.70 &48.95 &27.73 &29.08 &25.38 &15.41 &14.46\\
    TPVFormer~\cite{huang2023tri} & 27.83 &7.22 &38.90 &13.67 &40.78 &45.90 &17.23 &19.99 &18.85 &14.30 &26.69 &34.17 &55.65 &35.47 &37.55 &30.70 &19.40 &16.78\\ 
    OccFormer~\cite{zhang2023occformer} & 21.93 &5.94 &30.29 &12.32 &34.40 &39.17 &14.44 &16.45 &17.22 &9.27 &13.90 &26.36 &50.99 &30.96 &34.66 &22.73 &6.76 &6.97\\
    RenderOcc~\cite{pan2024renderocc} & 26.11 & 4.84 & 31.72 & 10.72 & 27.67 & 26.45 &13.87 &18.2 &17.67 &17.84 &21.19 &23.25 &63.2 &36.42 &46.21 &44.26 &19.58 & 20.72\\
    \midrule
    \ours (ours) & \textbf{28.45} & \textbf{9.52} & 24.15 & \textbf{15.85} & \textbf{34.36} & \textbf{34.62} & \textbf{16.29} & \textbf{17.37} & 15.25 & 12.04 & \textbf{27.03} & \textbf{26.91} & 68.66 & 35.41 & 45.13 & \textbf{45.33} & \textbf{27.72} & \textbf{28.09} \\
    \bottomrule
  \end{tabular}}
  \vspace{-5pt}
  \label{tab:nuscenes}
\end{table*}

\subsection{Loss Functions}


To train \ours, we use cross-entropy loss to supervise predicted 3D occupancy and rendered 2D semantic maps,  $L_1$ loss for rendered depths, and angular and $L_1$ losses for surface normal maps following~\cite{yu2022monosdf}. 
Distortion loss~\cite{barron2022mip} is also used to regularize the volume rendering weights. The final loss function is given as follows:
\begin{align}
    \mathcal{L}_{\text{total}} &= \mathcal{L}_{\text{sem}}(O, \hat{O}) + \lambda_{d}\mathcal{L}_{\text{dep}}(D, \hat{D})\notag\\
    &+\lambda_{s}\mathcal{L}_{\text{sem}}(S, \hat{S}) + \lambda_{n}\mathcal{L}_{\text{normal}}(N, \hat{N})+\lambda_{r}\mathcal{L}_{\text{reg}}(\tau),
\end{align}
where ${O}$, ${D}$, ${S}$, ${N}$ and $\tilde{O}$, $\tilde{D}$, $\tilde{S}$, $\tilde{N}$ are the predicted and ground-truth occupancy, depths, semantics, and normals, respectively and $\lambda_d, \lambda_s, \lambda_n, \lambda_r$ are the weights that balance the loss terms.

\begin{table}[t!]
  \small
  \centering
  \caption{\small Results on Occ3D-nuScenes validation set in terms of 3D reconstruction metrics.}
  \vspace{-5pt}
  \begin{tabular}{lcccc}
    \toprule
    \textbf{Method} & GT & Precision & Recall & F-score\\
    \midrule
    RenderOcc~\cite{pan2024renderocc} & \checkmark & 59.97 & 82.23 & 53.09 \\
    \midrule
    \ours (ours) & \checkmark & \textbf{67.21} & 76.98 & \textbf{71.80}\\
    \bottomrule
  \end{tabular}
  \vspace{-5pt}
  \label{tab:scenerec}
\end{table}

\section{Experiments}

\begin{figure*}[t!]
    \centering
    \includegraphics[width=0.98\textwidth]{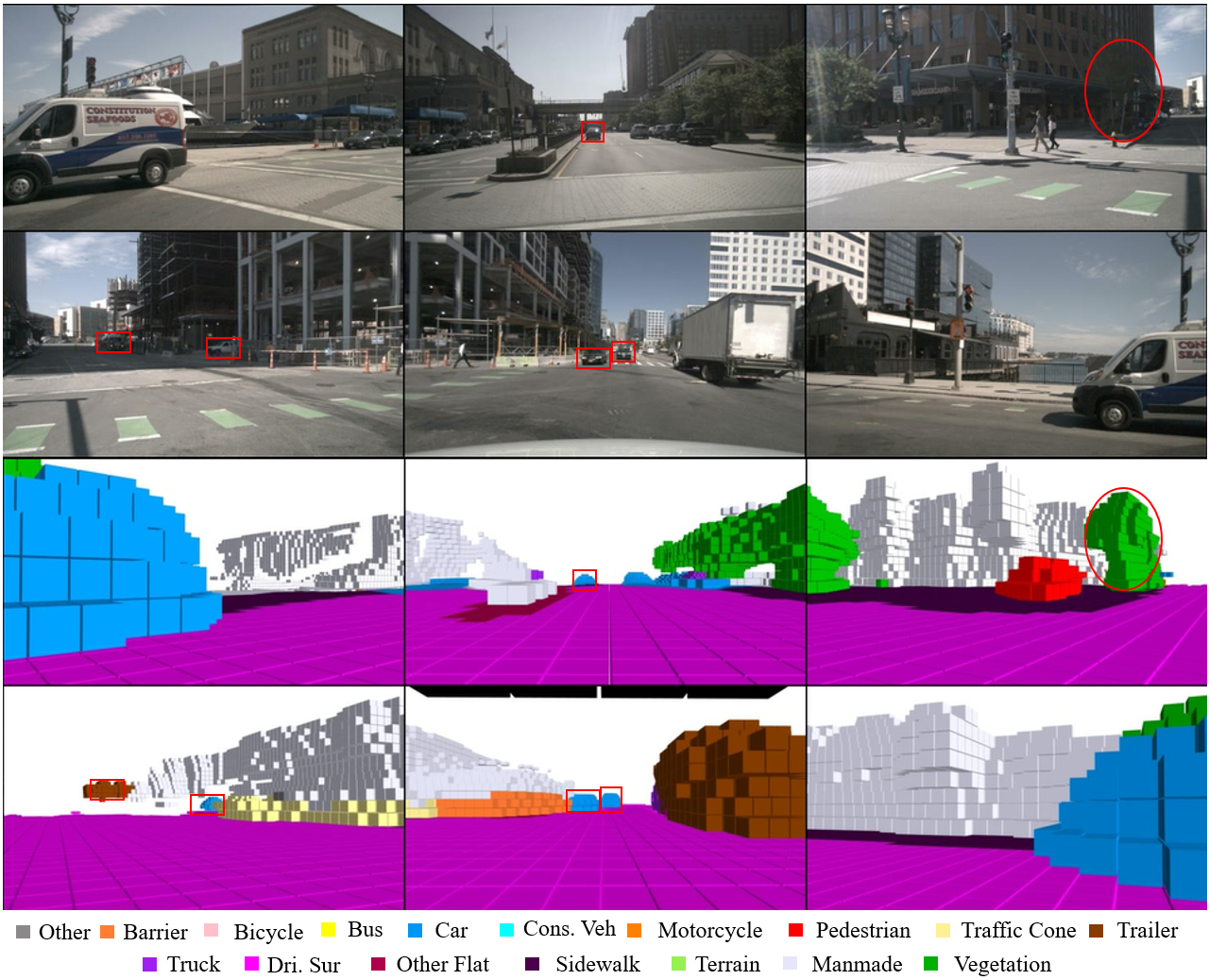}
    \vspace{-3pt}
    \caption{\small Example qualitative 3D semantic occupancy prediction of \ours on Occ3D-nuScenes validation set. Cons. Veh stands for ``Construction Vehicle'' and Dri. Sur stands for ``Drivable Surface''. We  see that \ours is able to capture objects  with fine details at far distances and obscure locations, as well as under poor lighting conditions, \eg, far-away vehicles, tree in a low-constrast region of the top-right image. This validates the benefit of introducing rich supervision into occupancy learning. For the back camera (middle of bottom row) we visualize the ego-car body at the top, avoiding obscuring drivable surface. Best viewed in color and zoomed in.}
    \label{fig:demo-res}
    \vspace{-7pt}
\end{figure*}

We evaluate \ours on Occ3D-nuScenes~\cite{caesar2020nuscenes,tian2024occ3d} and SemanticKITTI~\cite{behley2019semantickitti} benchmarks, and compare with latest state-of-the-art (SOTA) methods. We also conduct extensive ablation studies on  different design choices of \ours.

\subsection{Datasets and Setup}

\noindent\textbf{Occ3D-nuScenes}: nuScenes consists of 1,000 scenes captured by a synchronized camera array of 6 cameras. The dataset is split into 700 scenes for training, 150 scenes for validation and 150 scenes for testing. Occ3D-nuScenes boostraps nuScenes and annotates 3D semantic occupancy ground-truth consisting of 17 classes. The voxel grid range is $[-40\text{m}, -40\text{m}, -1\text{m}, 40\text{m}, 40\text{m}, 5.4\text{m}]$ along the $X,Y$ and $Z$ axis. The grid resolution is $200\times 200\times 16$ with a voxel size of $0.4$m. The input image resolution is $900\times 1600$.

\noindent\textbf{SemanticKITTI}: SemanticKITTI is curated based on KITTI-Odometry~\cite{geiger2012we,geiger2013vision}, which includes 22 sequences with a split of 10 sequences for training, 1 sequence for validation and 11 sequences for testing, respectively. It consists of 19 classes for the semantic occupancy ground-truth. The voxel range is $[0\text{m}, 51.2\text{m}, -25.6\text{m}, 25.6\text{m}, -2\text{m}, 4.4\text{m}]$, with grid resolution of $256\times 256\times 32$ along the $X, Y$ and $Z$ axis. The voxel size is $0.2$m. The input image resolution is $376\times 1241$. We use the RGB images from the left camera as input to our model.

We note that in this work we only consider the single-time instance setting, where the model consumes multi-camera images acquired at the same time instance, and do not use video frames as input. We compare with existing, non-video-based SOTA models that also use single-time instance camera inputs.

\subsection{Evaluation Metrics}

For evaluating 3D semantic occupancy prediction, we use the mean Intersection over Union (mIoU) of occupied voxels, averaging over all semantic classes
\begin{equation}
    \text{mIoU} = \frac{1}{C}\sum_{i=1}^C\frac{TP_i}{TP_i + FP_i + FN_i},\notag
\end{equation}
where $TP, FP, FN$ denotes the number of true positive, false positive and false negative predictions. To also evaluate the overall quality of learned scene geometry, we ignore semantic classes and make use of the precision, recall and F-score metrics commonly used in 3D reconstruction~\cite{bozic2021transformerfusion,murez2020atlas} 
\begin{align}
    \text{Prec} &=\frac{1}{|P|}\sum_{p\in P}\min_{p^{*}\in P^{*}}||p - p^{*}|| < \delta,\notag\\
    \text{Rec} &=\frac{1}{|P^{*}|}\sum_{p^{*}\in P^{*}}\min_{p\in P}||p - p^{*}|| < \delta,\notag\\
    \text{F-score} &=2\,/\,(1\,/\,\text{Prec} + 1\,/\,\text{Rec}),\notag
\end{align}
where $\delta$ is the threshold. 

\subsection{Implementation Details}

We use ResNet101~\cite{he2016deep} as the image backbone with pretrained weights from FCOS3D~\cite{wang2021fcos3d} to extract multi-camera image features. On Occ3D-nuScenes, we resize the input images to $352\times 704$. On SemanticKITTI, the input images are resized to $192\times 640$. We use AdamW~\cite{loshchilov2017decoupled} as the optimizer with an initial learning rate of $2\times 10^{-4}$. We train \ours with batch size $8$ for $24$ epochs for both datasets. We set weights $\lambda_d = \lambda_s = \lambda_n = 0.05$, while $\lambda_{\text{reg}}$ is set to $0.005$. Threshold $\delta$ is set to the same as voxel size on both datasets for evaluating 3D reconstruction quality.

\subsection{Main Results}
\vspace{-1pt}

\subsubsection*{Occ3D-nuScenes}


Table~\ref{tab:nuscenes} summarizes the 3D semantic occupancy results on Occ3D-nuScenes. \ours sets a new state-of-the-art performance with an mIoU of 28.45, outperforming existing SOTA methods including the very recent RenderOcc~\cite{pan2024renderocc}, which demonstrates the efficacy of our proposed approach. Note that while RenderOcc~\cite{pan2024renderocc} also leverages additional supervision during training, its auxiliary-ray scheme introduces misalignment. We employ a simple yet effective strategy by casting a ray for every pixel in the scene which ensures coverage. Fig.~\ref{fig:demo-res} provides qualitative examples of the 3D occupancy prediction by our proposed \ours. We see that \ours is able to capture objects at long distances and obscure locations with fine details, \eg, faraway cars. It is also robust under non-ideal lighting and contrast conditions, \eg, tree highlighted in top-right image. We attribute this to the auxiliary tasks that provide rich supervision into the system.

Since 3D reconstruction is also an important aspect of 3D occupancy prediction, which is useful for 3D scene understanding and downstream tasks, we additionally evaluate the 3D reconstruction quality using the standard metrics. Table~\ref{tab:scenerec} shows that our proposed \ours achieves a F-score of 71.80, significantly outperforming existing SOTA.


\begin{table*}[t!]
  \small
  \centering
  \caption{\small 3D Semantic Occupancy results on SemanticKITTI test set~\cite{behley2019semantickitti}.}
  \vspace{-5pt}
  \resizebox{\textwidth}{!}{
  \begin{tabular}{l|c|ccccccccccccccccccc}
    \toprule
    \textbf{Method} & \makecell{SSC\\mIoU} & \makecell{\begin{turn}{90}car\end{turn}} & \makecell{\begin{turn}{90}bicycle\end{turn}} & \makecell{\begin{turn}{90}motorcycle\end{turn}} & \makecell{\begin{turn}{90}truck\end{turn}} &  \makecell{\begin{turn}{90}other. veh\end{turn}}& \makecell{\begin{turn}{90}person\end{turn}} & \makecell{\begin{turn}{90}bicyclist\end{turn}} & \makecell{\begin{turn}{90}motorcyclist\end{turn}} & \makecell{\begin{turn}{90}road\end{turn}} & \makecell{\begin{turn}{90}parking\end{turn}} & \makecell{\begin{turn}{90}sidewalk\end{turn}} & \makecell{\begin{turn}{90}other. grnd\end{turn}}& \makecell{\begin{turn}{90}building\end{turn}} & \makecell{\begin{turn}{90}fence\end{turn}} & \makecell{\begin{turn}{90}vegetation \end{turn}} & \makecell{\begin{turn}{90}trunk\end{turn}} & \makecell{\begin{turn}{90}terrain\end{turn}} & \makecell{\begin{turn}{90}pole\end{turn}} & \makecell{\begin{turn}{90}traf. sign\end{turn}}\\
    \midrule
    AICNet~\cite{li2020anisotropic} & 6.73 & 15.30 &0.00 &0.00 &0.70 &0.00 &0.00 &0.00 &0.00 &39.30 &19.80 &18.30 &1.60 &9.60 &5.00 &9.60 &1.90 &13.50 &0.10 &0.00\\
    VoxFormer~\cite{li2023voxformer} & 12.35 &25.79 &0.59 &0.51 &5.63 &3.77 &1.78 &3.32 &0.00 &54.76 &15.50 &26.35 &0.70 &17.65 &7.64 &24.39 &5.08 &29.96 &7.11 &4.18\\
    TPVFormer~\cite{huang2023tri} & 11.26 & 19.20 & 1.00 & 0.50 & 3.70 & 2.30 & 1.10 & 2.40 & 0.30 & 55.10 & 27.40 & 27.20 & 6.50 & 14.80 & 11.00 & 13.90 & 2.60 & 20.40 & 2.90 & 1.50 \\
    SurroundOcc~\cite{wei2023surroundocc} & 11.86 & 20.60 & 1.60 & 1.20 & 1.40 & 4.40 & 1.40 & 2.00 & 0.10 & 56.90 & 30.20 & 28.30 & 6.80 & 15.20 & 11.30 & 14.90 & 3.40 & 19.30 & 3.90 & 2.40\\
    OccFormer~\cite{zhang2023occformer} & 12.32 & 21.60 &1.50 &1.70 &1.20 & 3.20 &2.20 & 1.10 & 0.20 & 55.90 &31.50 & 30.30 & 6.50 & 15.70 & 11.90 & 16.80 & 3.90 & 21.30 & 3.80 & 3.70\\
    RenderOcc~\cite{pan2024renderocc} & 12.87 &24.90 &0.37 &0.28 &6.03 &3.66 &1.91 &3.11 &0.00 &57.20 &16.11 &28.44 &0.91 &18.18 &9.10 &26.23 &4.87 &33.61 &6.24 &3.38\\
    \midrule
    \ours (ours) & \textbf{13.28} & \textbf{27.90} &\textbf{2.10} &1.40 &\textbf{6.10} &3.50 &\textbf{2.90} &\textbf{4.40} &0.10 &55.50 &22.60 &28.80 &2.40 &13.60 &9.40 &\textbf{30.10} &2.90 &28.90 &5.90 &3.90\\
    \bottomrule
  \end{tabular}}
  \vspace{-0pt}
  \label{tab:semkitti}
\end{table*}

\begin{table*}[t!]
  \small
  \centering
  \caption{\small Efficiency comparison. Measurements conducted on NVIDIA A100 GPU with PyTorch FP32. mIoU is on Occ3D-nuScenes. Note SurroundOcc is trained with its own curated ground-truth instead of Occ3D-nuScenes.}
  \vspace{-5pt}
  \begin{tabular}{p{3cm}p{2.5cm}p{2cm}p{2cm}p{2cm}p{2cm}}
    \toprule
    \textbf{Method} & Backbone & MACs & Latency & GPU Memory &mIoU \\
    \midrule
    BEVFormer~\cite{li2022bevformer} & Res101-DCN & 1.32T & 0.28s& \textbf{4.5G} & 23.67 \\
    TPVFormer~\cite{huang2023tri} & Res101-DCN & 1.29T & 0.29s & 5.1G & 27.83\\
    SurroundOcc~\cite{wei2023surroundocc} & Res101-DCN & 2.26T & 0.31s & 5.9G & -\\
    OccFormer~\cite{zhang2023occformer} & Res101-DCN& 2.63T & 0.29s & 9.4G & 21.93\\
    RenderOcc~\cite{pan2024renderocc} & Swin-base & 4.19T & 0.42s & 7.4G & 26.11\\
    \midrule
    \ours (ours) & Res101 & \textbf{0.77T} & \textbf{0.12s} & 4.9G & 28.45\\
    \bottomrule
  \end{tabular}
  \vspace{-5pt}
  \label{tab:efficiency}
\end{table*}

\subsubsection*{SemanticKITTI}

Table~\ref{tab:semkitti} shows the 3D semantic occupancy prediction results on the SemanticKITTI dataset. \ours sets a new SOTA performance with a mIou of 13.28. We note that since SemanticKITTI only features a single front-facing camera, the fact that \ours outperforms previous methods which require costly and complicated designs (\eg, 2D-3D spatial attention, auxiliary-ray sampling) further validates the effectiveness of our proposed rich, heterogeneous supervisions, as well as our simple yet effective per-pixel ray sampling strategy.

\subsubsection*{Computation Efficiency} Table~\ref{tab:efficiency} compares the various computation aspects of our proposed \ours method with other recent SOTA methods. One can see that \ours is significantly more efficient as compared to existing methods, with just 0.77 Tera-MACs of computation, which is more than 40\% and 81\% less costly as compared to TPVFormer and RenderOcc, respectively. When running on an NVIDIA A100 GPU, our \ours model has a significantly lower latency, over 57\% lower than all existing models while incurring a small GPU memory cost.

\begin{table}[t!]
  \small
  \centering
  \caption{\small Ablation study of different ray sampling frequency.}
  \vspace{-5pt}
    \begin{tabular}{lccc}
    \toprule
    \textbf{Method} & Strategy & Step size & mIoU\\
    \midrule
    RenderOcc~\cite{pan2024renderocc} & mip360 & \makecell{0.5*\text{voxel}\_\text{size}\\
    1.0*\text{voxel}\_\text{size}\\
    2.0*\text{voxel}\_\text{size}} &
    \makecell{26.11 \\ - \\ -}\\
    \midrule
    \ours (ours) & fixed & \makecell{0.5*\text{voxel}\_\text{size}\\
    1.0*\text{voxel}\_\text{size}\\
    2.0*\text{voxel}\_\text{size}}&
    \makecell{28.45 \\ 26.57 \\ 24.21}\\
    \bottomrule
  \end{tabular}
  \vspace{-10pt}
  \label{tab:raysampl}
\end{table}

\subsection{Ablation Study}

We conduct ablation studies on Occ3D-nuScenes on two aspects: ray sampling frequency in volume rendering, and using perspective supervisions in training.


\subsubsection*{Ray Sampling Frequency} Given that we employ per-pixel ray sampling, we experiment with different ray sampling frequencies (step size) in the volume rendering process when training our \ours model. As shown in Table~\ref{tab:raysampl}, increasing ray sampling frequency yields an improvement in the final semantic occupancy prediction performance. One can see that our per-pixel, fixed step-size sampling outperforms RenderOcc which adopts a more complicated sampling strategy introduced in mip-NeRF360~\cite{barron2022mip} even with fewer sampled points. This result further validates the design of \ours.

\subsubsection*{Supervision} Table~\ref{tab:supervision} shows the performance of \ours when training with only 3D labels, and training with both 3D labels and our proposed heterogeneous 2D supervisions. When trained only with 3D occupancy labels, the mIoU decreases to 26.04. However, this is still on par with the performance of other SOTA methods, indicating that our efficient 3D feature volume construction is effective. By further leveraging multi-task perspective supervisions including depth, semantics, and surface normal, \ours yields a considerable gain of mIoU at 28.45, demonstrating the benefit of our proposed learning scheme.

\begin{table}[t!]
  \small
  \centering
  \caption{\small Ablation study on supervision.}
  \vspace{-5pt}
  \begin{tabular}{lcc}
    \toprule
    \textbf{Method} & Supervision & mIoU\\
    \midrule
     \ours (ours) & \makecell{3D\\ 3D+2D} & \makecell{26.04 \\ 28.45}\\ 
    \bottomrule
  \end{tabular}
  \vspace{-10pt}
  \label{tab:supervision}
\end{table}

\section{Conclusion}

We propose H3O, a hyper-efficient and high performing system for vision-centric 3D semantic occupancy prediction. We construct an efficient 3D volume without costly attention-based 2D-3D transformation which significantly reduces the complexity of the entire system. To further improve 3D occupancy learning, we advocate the integration of auxiliary tasks such as multi-camera depth estimation, semantic segmentation and surface normal estimation through differentiable volume rendering, supervised by corresponding 2D labels. We also employ a simple yet effective ray sampling strategy. H3O achieves state-of-the-art performance on the Occ3D-nuScenes and SemanticKITTI benchmark, demonstrating the efficacy and superiority of our proposed method.







\newpage
\bibliographystyle{IEEEtran}
\bibliography{IEEEfull}

\end{document}